# A Self-Efficacy Theory-based Study on the Teachers' Readiness to Teach Artificial Intelligence in Public Schools in Sri Lanka

Chathura Rajapakse[1]

Department of Industrial Management, Faculty of Science, University of Kelaniya, Sri Lanka, chathura@kln.ac.lk

Wathsala Ariyarathna

Department of Information Technology, Faculty of Science and Technology, National Institute of Education, Sri Lanka, madurandi@gmail.com

Shanmugalingam Selvakan

Department of Information Technology, Faculty of Science and Technology, National Institute of Education, Sri Lanka, shanmugalingam@nie.edu.lk

**Objectives**. This paper explores teacher readiness for introducing artificial intelligence (AI) into Sri Lankan schools, drawing on self-efficacy theory. Similar to some other countries, Sri Lanka plans to integrate AI into the school curriculum soon. However, a key question remains: are teachers prepared to teach this advanced technical subject to schoolchildren? Assessing teacher readiness is crucial, as it is intricately linked to the overall success of this initiative and will inform the development of appropriate policies.

**Participants**. This study surveyed over 1,300 Sri Lankan public school teachers who teach Information and Communication Technology (ICT) using the snowball sampling approach. The respondents represent approximately 20% of the total ICT teacher population in Sri Lanka. Their readiness to teach AI was assessed using a general teacher self-efficacy scale specifically developed based on the well-established Self-Efficacy Theory. While key demographic factors like gender, education level, and educational background were also collected, their impact analysis is not included in this paper.

**Study Method**. The study was conducted based on the premise that teachers' readiness to teach AI hinges on their self-efficacy towards teaching AI in the classroom. This premise was substantiated through a review of existing research, and a conceptual model of teachers' self-efficacy for AI instruction was developed. To assess this model, a nationwide survey targeting school ICT teachers was conducted. The questionnaire used in the survey was based on existing research on evaluating teacher self-efficacy. Data analysis, focusing on testing and validating the conceptual model, primarily employed the PLS-SEM approach.

**Findings**. This study identified several key findings: 1) Teachers generally reported low self-efficacy regarding their ability to teach AI, 2) Teachers' self-efficacy was most influenced by their emotional and physiological states, as well as their imaginary experiences related to teaching AI, 3) Surprisingly, mastery experiences had a lesser impact on their self-efficacy for teaching AI, and 4) Neither vicarious experiences (observing others teach AI) nor verbal persuasion had a significant impact on teachers' self-efficacy. Additionally, the study revealed that the teachers' own level of expertise in AI, along with their social capital, is insufficient to deliver a standard AI curriculum.

---

[1] URL: https://dl.acm.org/doi/abs/10.1145/3691354 DOI: https://doi.org/10.1145/3691354

**Conclusions**. The analysis of the results found that Sri Lankan teachers currently lack the readiness to teach AI in schools effectively. Potential lapses in certain sources of self-efficacy were also identified. It further revealed the need for a more systemic approach to teacher professional development. Therefore, the study recommends further research exploring the potential of incorporating a socio-technical systems perspective into the government's teacher training initiatives.

# 1 Introduction

The need for and challenges of teaching artificial intelligence (AI) at primary, secondary, and upper-secondary levels have been a major focus of recent academic discussions [1],[2],[3]. Often referred to as AI4K12 [4], this area explores global initiatives that introduce AI to students from kindergarten through high school. The rapid advancements in deep learning and generative AI technologies suggest AI will become a transformative force. This realisation has prompted governments and policymakers to recognise the need to prepare future citizens for a world heavily influenced by AI. As AI becomes increasingly integrated into information systems, concerns are mounting about citizens' ability to use these systems responsibly and understand the consequences of not doing so [5]. Furthermore, anxieties regarding AI's potential impact on societal sustainability highlight the need to equip future workforces with the skills to combine human creativity with AI's potential to create sustainable systems. In light of these issues, many countries worldwide recognise the critical importance of educating future generations about artificial intelligence from a young age [6].

However, integrating artificial intelligence (AI) into K-12 education presents a complex challenge [7]. Effectively implementing AI4K12 initiatives requires addressing a multifaceted landscape that encompasses educational standards, curriculum design, formal and informal learning opportunities, student learning outcomes, teacher professional development, and the continuity of the learning process (i.e., learning progression) [8]. As AI education in K-12 settings is an emerging field, numerous research gaps need to be filled to optimise AI4K12 programs. Key unanswered questions include: 1) Policy and Guidelines: What types of policies and guidelines should be established [4],[9],[10],[11]? 2) Content Design: What kind of content is most suitable for different age groups [12],[13]? 3) Delivery Methods: What is the most effective approach for delivering AI education [14],[15],[16]? 4) Learning Environment: How can we leverage both formal and informal education to enhance students' AI competency [17],[18],[19]? 5) Assessment Strategies: How should we assess student learning outcomes related to AI [8],[20],[21]? 6) Teacher Preparedness: Are teachers adequately equipped to deliver AI-related content [22],[23],[24]? 7) Teacher Confidence: How can we bolster teacher confidence in teaching AI topics [8],[25]? 8) Learning Continuity: How can we ensure a continuous and progressive AI learning experience for students [8],[18],[19]? The global education landscape is characterised by significant diversity in social norms, culture, language, economy, and digital access. A successful AI4K12 initiative in one country may not translate effectively to another. Therefore, further research from diverse regions is crucial to inform policymakers' decisions regarding AI4K12 implementation.

Sri Lanka has established itself as a leading destination for offshore software development outsourcing for several decades. This success can be largely attributed to the country's high-quality education system, which has produced a skilled workforce serving key markets like Australia, the UK, the USA, and others [26]. Notably, Sri Lanka boasts a 71% computer literacy rate among those completing secondary education, a testament to the effectiveness of its ICT education programs despite a significant digital divide within the country [27]. Recognising the evolving needs of the global economy, the Sri Lankan government's proposals for general education reforms (2022-2032) emphasise two key objectives: 1) fostering a workforce that can leverage science and technology to drive innovation amidst rapid technological advancements and 2) developing a skilled population capable of transforming key sectors like agriculture, manufacturing, and tourism into pillars of the national economy [28]. In the era of Education 4.0, which mirrors Industry 4.0, equipping future generations with artificial intelligence (AI) skills is crucial to achieving these goals [29]. Consequently, the Sri Lankan government has made a significant move by integrating AI into the school ICT curriculum for grades 8 and above, starting in 2024 [30]. This decision aligns perfectly with the strategic objectives outlined by SLASSCOM, the representative body for Sri Lanka's IT/BPM industry, and is poised to propel further the nation's IT/ITeS sector [26].

## 2 Problem

Empowering teachers is a crucial aspect of integrating AI into school ICT education [31],[32]. Effectively delivering a curriculum centred on core AI concepts like perception, knowledge representation and reasoning, learning, natural interactions, and societal impact [33] requires teachers to possess a broader understanding of AI technologies, pedagogical knowledge, and strong mental preparedness [34]. A significant challenge lies in the fact that most teachers are not digital natives, unlike their students. This can make teaching an advanced subject like AI particularly demanding [35]. Research indicates that many teachers express concerns about teaching AI and report a lack of confidence due to limited access to relevant knowledge in technology, pedagogy, and content [8]. The authors' own interactions with Sri Lankan ICT teachers over several years confirm this observation. Despite the need for teacher preparation, there's a scarcity of studies assessing teacher readiness for AI4K12 initiatives in various countries. Since Sri Lanka has chosen to incorporate AI into its school ICT curriculum, evaluating the current teacher workforce's preparedness to teach this subject is essential. Such an evaluation would inform policymakers about the necessary teacher professional development approaches to enhance their confidence and equip them to support the AI4K12 initiative effectively. Thus, the two critical questions addressed in this study are:

- How ready is the Sri Lankan ICT teacher workforce to teach AI at the K12 level?
- What teacher professional development approach would make them better prepared and confident to teach the subject?

### 2.1 Review of Relevant Scholarship

Research on integrating AI into K-12 education has grown significantly in recent years. This is particularly evident in the surge of published literature between 2020 and 2023. This trend highlights the field's novelty and the growing global interest in this field of study. However, the majority of current research focuses on curriculum development and teaching methods (pedagogy). Additionally, the expanding body of literature on STEAM education increasingly explores the effective use of AI to introduce STEM subjects at the K-12 level [36],[37]. Broadly, the connection between AI and education can be categorised into three areas: (1) learning with AI, (2) learning about AI, and (3) preparing students for a future influenced by AI [38]. This study specifically excludes the first category (learning with AI) and focuses on the latter two. In simpler terms, our research targets the readiness of teachers who will directly teach AI, not other STEM instructors who might leverage AI to enhance their own subject delivery.

As previously mentioned, relatively few researchers have explored teachers' perspectives on teaching AI in K-12 education. One notable study by Yau et al. examined teachers' conceptions of delivering AI in K-12 schools [39]. They emphasise the importance of understanding these conceptions when developing professional development programs for teachers. They argue that the variations in teachers' views on AI education can reveal their strengths and weaknesses in teaching AI, inform educators and policymakers on how to enhance teachers' AI competence, and ultimately contribute to the advancement of general AI education. This phenomenological study identified six key teacher conceptions: technology bridging, knowledge delivery, interest stimulation, ethics establishment, capability cultivation, and intellectual development. Interestingly, most of the participants in this study were ICT teachers with an ICT-related educational background. This background might explain their generally positive perception (potentially indicating greater readiness) towards AI education for K-12 students. Furthermore, African high school computer science teachers' preconceptions about teaching machine learning were investigated by Sanusi et al. [40]. The phenomenographic analysis involving twelve African in-service computer science teachers revealed five categories of preconceptions, namely supporting student technical knowledge, having knowledge of the concept, focusing on professional development practices, contextualising teaching resources and tools, and sustainability for development goals. A noteworthy finding from the study is the teachers' lack of knowledge and training to teach machine learning in schools. The study emphasises the need for professional development training for teachers to teach machine learning to promote better learning outcomes.

Polak et al. [31] utilised the "will, skills, and tools" model to explore the perspectives of teachers from four European countries (Bulgaria, Greece, Italy, and Romania) on integrating digital competencies for AI into their students' learning. They emphasise the critical role teachers play in fostering classroom innovation and developing novel educational approaches for AI education. Consequently, understanding teacher attitudes and perceptions towards AI education is essential. The study reveals a very positive teacher attitude (strong will) towards AI education, but it also identifies a lack of competence in using AI tools (inadequate skills). Based on these findings, the authors recommend prioritising the development of teachers' skills to enhance their support for AI education. In another focused group study that involves Swedish in-service teachers, Velander et al. report a lack of understanding of AI and its limitations, as well as emotional responses related to participants' preconceptions [32]. However, it is noteworthy that both of these studies involve not only the ICT teachers but also the teachers from other disciplines.

Another relevant study by Ayanwale et al. [41] closely aligns with the present research and offers a comprehensive examination of teacher readiness for K-12 AI education. They emphasise, along with other researchers, the importance of assessing teacher readiness and providing professional development opportunities. Their work highlights that while professional development is crucial, understanding teachers' baseline intention and readiness to teach AI is equally important. After all, teachers' acceptance and enthusiasm for teaching new technologies can significantly impact their teaching practices. Ayanwale et al. ground their research in the Theory of Planned Behavior (TPB) [42]. They employ the variance-based structural equation modelling approach to explore the relationships between eight key variables, namely AI anxiety, Perceived usefulness of AI, AI for social good, Attitude towards using AI, Perceived confidence in teaching AI, Relevance of AI in K-12 education, AI readiness, and Behavioural intention to teach AI. Their findings reveal that confidence in teaching AI directly predicts a stronger intention to teach AI, while the perceived relevance of AI has a powerful influence on teacher readiness. Notably, the study defines behavioural intention as a measure of teachers' willingness to teach AI. In simpler terms, their results suggest that higher perceived confidence translates to a greater willingness to teach AI. On the other hand, teacher readiness is defined as the degree to which an individual feels equipped to deliver AI education at the K-12 level. According to the study, a stronger perception of AI's relevance in K-12 education leads to higher teacher readiness. Additionally, the research highlights significant positive relationships between teacher readiness and factors like confidence and the perception of AI as a force for social good. A strong correlation between behavioural intention and readiness was also found.

Beyond the field of K-12 education (AI4K12), Badiah N. M. Alnasib has also investigated factors affecting faculty members' willingness to integrate artificial intelligence (AI) into their teaching practices within Saudi Arabia [43]. This study appears to be grounded in the theory of planned behaviour, as the researchers developed a conceptual model that hypothesises readiness to integrate AI is influenced by perceived benefits of AI in higher education, perceived benefits of AI in teaching, available facilities and resources, behavioural intention, and attitude towards AI. Their findings indicate statistically significant correlations at the 0.01 level between faculty members' readiness to incorporate AI into teaching and the perceived benefits of AI in higher education and teaching, attitudes towards AI, intentions to use AI, and the presence of conditions that facilitate AI use.

The importance of studying teachers' preconceptions and preparedness for teaching AI in schools is well-established in the existing literature. However, research in this area, particularly across different regions of the world, remains limited. This is significant because the diversity of the teaching workforce and potential disparities in resource distribution between countries can lead to vastly different outcomes for AI-in-K12 (AI4K12) initiatives. This gap highlights the need for more research on teacher readiness in various regions implementing AI4K12 programs. Notably, despite Sri Lanka's rapid rollout of AI education from lower secondary levels upwards, no studies have been conducted to explore teacher readiness in this context. The present study aims to address this critical research gap.

## 2.2 Hypothesis, Aims, and Objectives

This study investigates the readiness of Sri Lankan public school teachers to teach AI. The goal is to inform administrators in designing effective professional development programs that can enhance teacher confidence in delivering AI education. The study utilises self-efficacy theory [44] as its theoretical framework, differing from the theory of planned behaviour (TPB) used in some related research. This choice is justified because the obligation to teach AI in Sri Lankan public schools stems from a government policy, making it a mandatory requirement rather than a planned behaviour for teachers. In other words, teachers are expected to cover the AI content in the government-approved curriculum regardless of their personal preferences. Self-efficacy theory is a more suitable lens through which to examine teacher readiness. Readiness, defined as a teacher's confidence in delivering AI education to students from elementary to high school [41], aligns well with the concept of self-efficacy. Teacher self-efficacy, which refers to a teacher's belief in their ability to impact student learning outcomes [45], reflects a teacher's willful self-confidence (and thereby the readiness) to teach this complex subject in the classroom. Several studies have explored the connection between teacher readiness and self-efficacy. Şatgeldi's master's thesis on developing an instrument for science teachers' perceived readiness for STEM education highlights self-efficacy as a key determinant of readiness [46]. Citing Baker [47], Şatgeldi suggests that teachers with low self-efficacy may shy away from teaching concepts they find difficult. Another study by Baker examines the relationship between teachers' self-efficacy regarding general classroom management skills and their readiness (defined as both ability and willingness) to adapt specific behaviour management techniques to individual student needs [48]. Baker further investigates the link between teachers' beliefs about their interpersonal self-efficacy for classroom management and their readiness to implement targeted behaviour management strategies [47]. Integrating AI into the curriculum can also be viewed as a desired organisational change. Cunningham suggests that employees with higher self-efficacy towards change and who actively contribute to change initiatives can facilitate smoother implementation [49]. Similarly, Bagus Emsza et al.'s study on the relationship between self-efficacy and change readiness found that employees with high self-efficacy are more receptive to briefings and empowerment, leading to more effective change implementation [50]. Their findings support the argument that teacher self-efficacy significantly impacts their readiness to adopt the desired change of teaching AI as part of the school ICT curriculum.

Bandura defined self-efficacy as "beliefs in one's capabilities to organise and execute the courses of action required to produce given attainments" [51]. He identified four key factors that influence self-efficacy: mastery experience obtained by successfully performing a challenging task, vicarious experience obtained by seeing others perform threatening activities without adverse consequences, verbal persuasion obtained through positive (or negative) feedback, and physiological and emotional states reached through the arousal of emotions confirming (or questioning) one's competence of delivering the particular task. Teacher efficacy has been widely studied and has been consistently found to be associated with teaching performance and students' learning achievements [52],[53]. In the context of teacher self-efficacy, mastery experiences refer to the sense of accomplishment gained through effective teaching in both routine and challenging situations. Vicarious experiences involve observing other teachers from the same subject area successfully complete similar tasks. Verbal persuasion refers to the development of self-efficacy through positive or negative feedback received on teaching practice. Physiological and emotional states encompass the level of arousal a teacher experiences while teaching, which can influence their self-perceptions of competence. Among these factors, research suggests that mastery experience and verbal persuasion are the most influential in shaping teacher self-efficacy [52],[53]. Building on Bandura's self-efficacy model, Maddux and Gosselin propose a fifth source: imaginary experiences [54]. They define imaginary experiences as individuals' mental simulations of themselves or others succeeding or failing in hypothetical situations. In the context of teaching, this could involve visualising oneself performing effectively as a teacher in a classroom setting.

The experiments in this paper are designed based on this five-factor model of self-efficacy formation. Figure 2.1 illustrates the initial conceptual model adopted for this study.

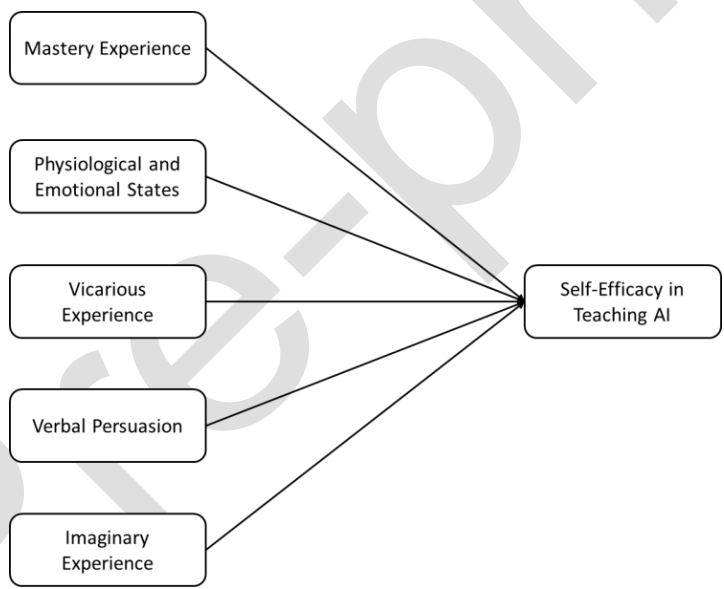

**Figure 2.1:** **The Conceptual Model on Teachers' Readiness to Teach Artificial Intelligence**

The primary hypotheses initiated for this study are as follows.

H1: There is a positive correlation between *Mastery Experience* and *Self-efficacy in Teaching AI*

H2: There is a positive correlation between Physiological and Emotional States and Self-efficacy in Teaching AI

H3: There is a positive correlation between *Vicarious Experience* and *Self-efficacy in Teaching AI*

H4: There is a positive correlation between *Verbal Persuasion* and *Self-efficacy in Teaching AI*

H5: There is a positive correlation between *Imaginary Experience* and *Self-efficacy in Teaching AI*

## 3 Method

### 3.1 Sampling, Distribution, and Ethnic Diversity

This study targeted in-service public-school teachers in Sri Lanka who teach Information and Communication Technology (ICT). Sri Lanka has a mature ICT teacher workforce of 6,500 with extensive experience delivering a continuously updated ICT curriculum since 2007. As a national policy, ICT is a mandatory subject from grade 6 onwards, aligning with international secondary and upper-secondary levels. Following a curriculum revision in 2021-2022, the Sri Lankan government plans to integrate Artificial Intelligence into the school ICT curriculum starting in 2024. Therefore, current in-service ICT teachers represent the most relevant population for this study. Furthermore, the Sri Lankan ICT curriculum is typically delivered in the two national languages, Sinhala and Tamil, as well as English. This necessitates including teachers from both Sinhalese and Tamil ethnicities in the respondent pool.

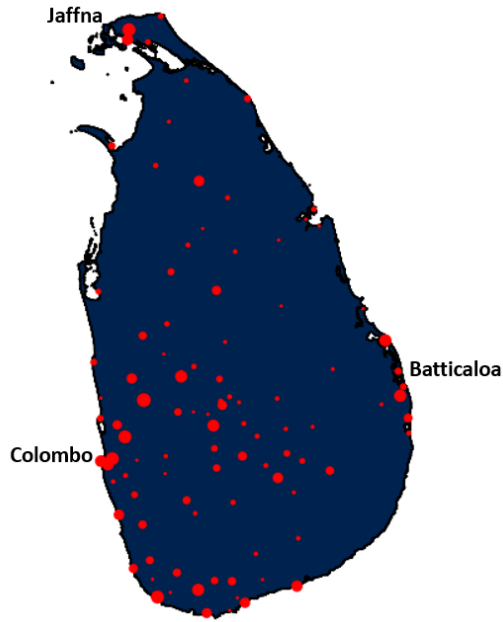

**Figure 3.1:** **The geographical distribution of the respondents**

This study employed a snowball sampling method, leveraging authors' connections with various government education institutions (Ministry of Education, National Institute of Education) involved in education policy, teacher professional development and curriculum development. This facilitated reaching a diverse initial sample of respondents across ethnicities and geographical locations. These initial respondents were then asked to share the questionnaire with others in their professional networks. The questionnaire was prepared electronically for easy distribution via email and social media, maximising potential reach. The goal of snowball sampling was to gather a large number of responses with good geographical representation. To accommodate respondent preferences, the questionnaire was available in both Sinhala and Tamil. Careful attention was given when selecting the initial respondents to mitigate potential bias in the snowball sampling method.

The questionnaire was administered in June and July 2023, yielding 1236 responses. This represents roughly 20% of the target population. Figure 3.1 illustrates the geographical distribution of respondents. Notably, responses were received from all populated areas of the country, with a sufficient concentration in all populated areas of the country. Furthermore, the data reflects the ethnic diversity of the country with substantial representation from Tamil-speaking regions like Nothern, Central, and Eastern provinces. The Muslim-majority areas in the Eastern province are also well-represented.

## 3.2 Constructs, Measures and Scale Development.

The conceptual model in Figure 2.1 identifies five factors influencing teacher self-efficacy: mastery experience, physiological and emotional states, vicarious experience, verbal persuasion, and imaginary experience. These factors are considered exogenous latent constructs (independent variables), while self-efficacy is the endogenous latent construct (dependent variable). To measure these constructs, a scale was developed based on existing teacher self-efficacy measurement scales and relevant literature [55]. While numerous researchers have attempted to create methods for measuring teacher efficacy [55], challenges arise due to conceptual ambiguity surrounding the construct [56]. Several prominent self-efficacy questionnaires exist, including the generalised self-efficacy scale [57], teacher self-efficacy scale [58], and science teachers' self-efficacy scale [59]. However, due to its clarity and recent development, the Sources of Teacher Efficacy Questionnaire (STEQ) (by Hoi et al. [52] was chosen as the baseline scale for this study.

Hoi et al. developed the STEQ with two goals: to create a scale for assessing teacher self-efficacy and to examine the relationship between self-efficacy and its four main sources, as identified by Bandura: Mastery Experience, Vicarious Experience, Physiological and Emotional State, and Verbal Persuasion. The STEQ's 26 items were primarily adapted, with some revisions, from the Sources of Middle School Mathematics Self-Efficacy Scale by Usher and Pajares [60]. To ensure clarity, accuracy, and compliance, six experienced teachers reviewed these items. Additionally, a second scale, the Teachers' Sense of Efficacy Scale (TSES) [55], was adopted to measure the respondents' level of self-efficacy (the dependent variable) separately. The TSES, previously established as a three-factor model (instructional strategies, classroom management, and student engagement), has demonstrated reliability and validity across various populations [55]. Prior to evaluating the STEQ's reliability and validity, the newly collected data was used to re-assess the internal consistency and factorial structure of the TSES. The internal consistency of the three subscales (0.88 to 0.90) was satisfactory. Fit indices like the Comparative Fit Index (CFI), Tucker-Lewis Index (TLI), Root Mean Square Error of Approximation (RMSEA), and Standardized Root Mean Square Residual (SRMR) also indicated a good fit for the factorial structure. However, the researchers ultimately chose a second-order model for the factorial structure due to its superior fit compared to the three-factor model. The composite self-efficacy score derived from this second-order TSES model served as the study's independent variable

According to Hoi et al., the STEQ demonstrates strong internal consistency, reliability, and validity. Confirmatory factor analysis was used to evaluate the STEQ's factorial structure. The four-factor model exhibited superior fit compared to alternative one-factor and second-order models ($\chi^2$ (293) = 636.40, SRMR = .057, RMSEA = .069 [.061, .076], TLI = .90, CFI = .91). Additionally, all standardised residual covariances were below an absolute value of 0.3, further supporting the four-factor model as a good fit for the data. Internal consistency reliability (alpha) was satisfactory for all four subscales: mastery experience (0.89), vicarious experience (0.86), social persuasion (0.90), and physiological arousal (0.87). The study also provides evidence for criterion-related validity. The four STEQ constructs exhibited moderate to high positive correlations (ranging from 0.63 to 0.79) with the composite self-efficacy factor derived from the TSES. Furthermore, a multiple regression analysis revealed that all four STEQ factors were significant predictors of self-efficacy (F = 146.94, p < 0.001). Therefore, although the scale developed for this study has not undergone formal reliability and validity testing, a strong foundation exists.

The self-efficacy scale items referenced in [52] are statements suitable for evaluation using a Likert scale. These statements were organised around the four core sources of self-efficacy, excluding imaginary experience. To ensure an accurate assessment of self-efficacy for teaching AI, each category's statements were carefully reviewed to select the most relevant ones. For online clarity, some statements were expanded. For example, the mastery experience statement "I have what it takes to succeed in teaching" was broken down into four statements reflecting subject knowledge, AI experience, ability to recognise common AI techniques, and continuous curiosity about emerging AI technologies. Similarly, statements that looked redundant were merged, and two new statements were added based on existing literature to assess imaginary experience, which was absent in the baseline scale. Moreover, three statements were added based on the literature to reflect the endogenous construct. A list of 37 initial statements was ultimately reduced to a more manageable list of twenty-five statements. A four-point Likert scale was adopted for each sentence, similar to the generalised self-efficacy scale [57], in which 1 implied 'Strongly Disagree' and 4 implied 'Strongly Agree'.

Although three items were added aiming to capture the endogenous construct, their reliability was not formally assessed. Studies like [52] typically employ a separate, established scale (e.g., TSES) to obtain a composite self-efficacy score when it serves as the independent variable. However, for the present island-wide online survey, administering a separate lengthy scale alongside the existing measures was impractical. As an alternative, the three items were chosen based on the theoretical framework presented by Hoy et al. [56]. They define teacher self-efficacy as the belief in one's capability to achieve desired student learning outcomes, even with challenging students. Furthermore, self-efficacy beliefs are thought to influence a teacher's effort, aspirations, persistence during difficulties, and resilience in the face of setbacks [56]. While the items were grounded in theory, the lack of formal reliability testing is a potential limitation. However, the alignment of the items with established theory suggests they likely captured self-efficacy effectively. Future research employing these items would benefit from a formal reliability assessment to strengthen the overall measurement strategy.

Additionally, the scale underwent a pilot study involving five experienced teachers. As highlighted by Gosselin et al. [54], gathering information through interviews and surveys with relevant individuals is crucial for constructing effective self-efficacy measures. This pilot served as a valuable refinement process, enhancing the scale's overall reliability and validity. Based on teacher feedback during the pilot study, the wording of statements was adjusted (for clarity), and a new construct – External Factors – was incorporated into the conceptual model. Teachers emphasised the significance of training, resources and the overall teaching environment provided by the government and school authorities, which ultimately impacts their confidence in teaching a challenging technical subject. Four additional statements were added to the scale to assess this new construct, resulting in a final version with 29 items. It is important to acknowledge that these four statements have limitations. They haven't been formally tested for reliability, nor are they directly supported by existing literature or aligned with established sources of self-efficacy. However, given their relevance based on teacher feedback from the pilot survey, the decision was made to include them despite these limitations. Figure 3.2 illustrates the updated conceptual model.

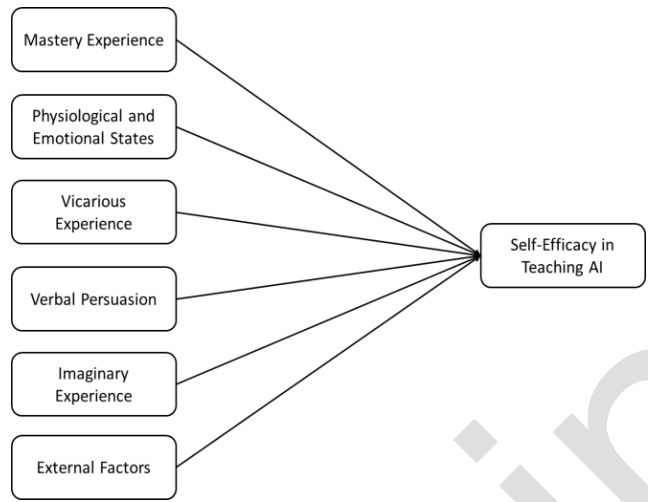

**Figure 3.2:** Finalized conceptual model used for the study

Accordingly, an additional hypothesis was added as follows.

H6: There is a positive relationship between the *External Factors* and the *Self-efficacy on Teaching AI.*

Table 3.1 displays the reflective statements used to assess self-efficacy. Each statement can be rated on a four-point Likert scale, with response options labelled (a), (b), (c), and (d) corresponding to point values of 10, 20, 30, and 40, respectively [57]. Statements prefixed with "EF" in the ID code indicate they are related to self-efficacy.

**Table 3.1:** Statements that assess Self-efficacy in teaching AI

| ID | Efficacy Question | (a) | (b) | (c) | (d) |
|---|---|---|---|---|---|
| EF01 | Teaching AI to school children can be challenging, but I am confident that I can overcome these challenges with hard work and dedication | | | | |
| EF02 | I am eager to get an opportunity to teach Artificial Intelligence | | | | |
| EF03 | When learning or teaching subjects related to new technology, I feel motivated and energetic | | | | |

Table 3.2 lists thirteen questions targeting mastery experience, a major source of self-efficacy. The remaining questions, categorised by their corresponding self-efficacy source, are presented in Table 3.3. As previously mentioned, question IDs use a prefix to denote the self-efficacy source they relate to. For instance, the prefix "MS" indicates a question about mastery experience.

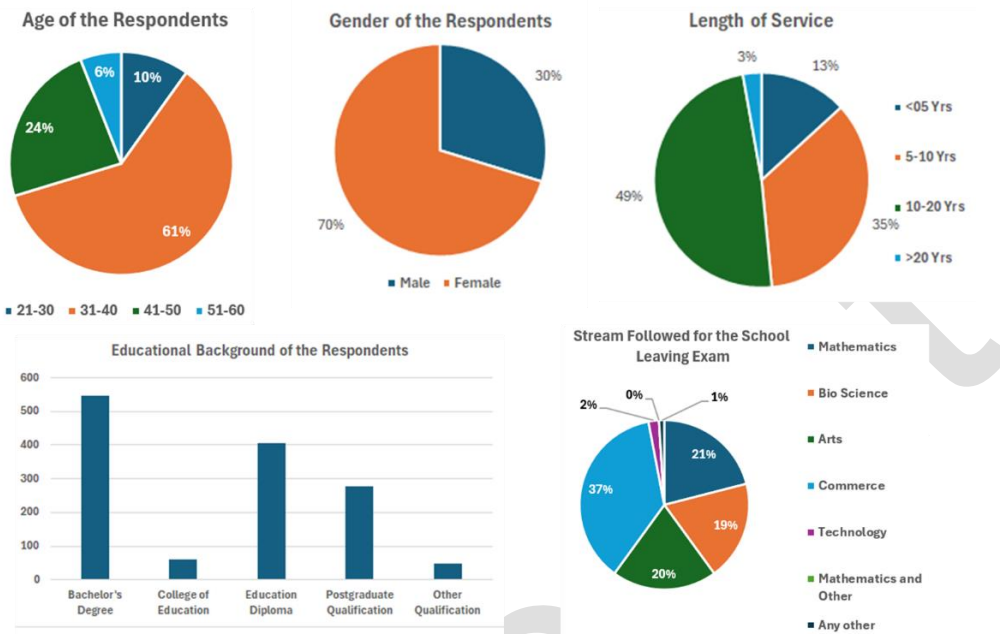

**Figure 3.3:** Key characteristics of the study participants

## 3.3 Data Collection

Data collection for this study utilised a structured questionnaire. In addition to statements assessing self-efficacy, the questionnaire gathered basic demographic information from respondents, including gender, educational background, school location, age group, and teaching experience. It also included a questionnaire item for respondents to self-rank their knowledge level on key topics in artificial intelligence literature. Throughout the process, anonymity was maintained – no data was collected that could identify individual participants. Of the initial 1236 responses, 116 were excluded due to missing data. This resulted in a final sample size of 1120 responses used for analysis. A summary of the sample demographics is provided below. The key characteristics of the study participants are presented in Figure 3.3.

- 70% of the respondents were females
- 60% of the respondents belonged to the age category of 31-40 years
- Only about 6% of the respondents were above 50 years old
- 44% had a bachelor's degree, and 22% had postgraduate qualifications
- Only 42% had a science education background
- Only 20% had studied mathematics for their school-leaving examination
- 50% of the respondents had a service of 10-20 years

The summary statistics suggest a strong representation of the Sri Lankan ICT teacher population in terms of age and gender. The comparatively younger teacher workforce is likely due to the recent introduction of ICT as a subject in the Sri Lankan curriculum (in 2007). It's also noteworthy that only about 40% of the teachers have a science education background. In other words, approximately 60% come from non-science backgrounds. This is a significant finding, as teaching a technical subject like AI ideally requires a scientific background from educators. According to a recent report by the European Digital Education Hub's AI in education squad, skills in computational thinking, specific AI topics, and mathematics are crucial for educators teaching AI [61]. Therefore, the majority of teachers having a non-science background could indicate less overall preparedness within the workforce to teach AI in schools.

**Table 3.2:** Statements that evaluate mastery experience

| ID | Efficacy Question | (a) | (b) | (c) | (d) |
|---|---|---|---|---|---|
| MS01 | At what level have you studied Artificial Intelligence (AI) at a university or National College of Education (NCoE)? | | | | |
| MS02 | I am confident in my ability to identify common applications of Artificial Intelligence (AI) and the techniques used to implement them | | | | |
| MS03 | I have applied artificial intelligence (AI) techniques to develop solutions for real-world problems. | | | | |
| MS04 | Learning and teaching new technologies has always presented challenges for me, but I have consistently been able to overcome them | | | | |
| MS05 | I actively stay informed about emerging artificial intelligence (AI) technologies by following their development and learning about them. This allows me to maintain a strong understanding of current AI advancements | | | | |
| MS06 | I frequently help friends, students, and colleagues understand artificial intelligence (AI) technologies by answering their questions effectively | | | | |
| MS07 | I actively seek out new teaching/learning tools, methods, and evaluation strategies related to artificial intelligence (AI), and integrate them into my teaching or learning practices | | | | |
| MS08 | While teaching Information Technology (IT) to young children who are comfortable with technology, I rarely encounter situations where my own knowledge is challenged, which can sometimes make me feel uneasy | | | | |
| MS09 | Students who take my Information Technology class typically achieve high scores on exams | | | | |
| MS10 | I make a consistent effort to attend seminars and workshops on Artificial Intelligence (AI) offered by various organizations and universities whenever possible | | | | |
| MS11 | In my teaching, I frequently encourage and guide students to explore innovative projects that utilize modern information technologies, including Artificial Intelligence (AI) | | | | |
| MS12 | I actively participate in media programs and workshops as a resource person to increase public awareness and understanding of modern information technologies | | | | |
| MS13 | I have cultivated a strong network of contacts within software development companies and other businesses that are actively using or exploring Artificial Intelligence (AI) | | | | |

# 4 Results

## 4.1 Measuring Overall Self-Efficacy

Self-efficacy was initially measured using the method developed by Schwarzer and Jerusalem in their generalised self-efficacy scale [57]. Each question related to self-efficacy offered four answer choices (a, b, c, and d) and assigned scores of 10, 20, 30, and 40 points, respectively. For example, if a respondent selected option (b) for a question, their answer would be scored as 20. An individual's overall self-efficacy score was then calculated by summing the scores from all self-efficacy questions. Figure 4.1 illustrates the distribution of these scores. The average self-efficacy score across all respondents was 647. However, when normalised using a min-max scaling approach, this value translates to 46%, suggesting a somewhat lower overall level of self-efficacy among the teachers. Interestingly, the indicator variables (EF01, EF02, EF03) representing the exogenous variable "Self-Efficacy" generally received higher scores. This inconsistency warrants further investigation of the overall self-efficacy score before drawing conclusions about the teachers' readiness to teach artificial intelligence.

**Table 3.3:** Statements that assess the remaining sources of self-efficacy

| ID | Efficacy Question | (a) | (b) | (c) | (d) |
|---|---|---|---|---|---|
| VC01 | I actively follow and am inspired by excellent teachers who excel at explaining complex subjects like Artificial Intelligence (AI) in a clear and engaging manner, tailored to the age and understanding level of their students. This approach helps me refine my own teaching style. | | | | |
| VC02 | My fellow Information and Communication Technology (ICT) teachers, with whom I regularly interact, share a strong interest in exploring and teaching Artificial Intelligence (AI). | | | | |
| VC03 | I have access to a mentor with expertise in artificial intelligence (AI) who can provide guidance and support in resolving any challenges I encounter while teaching or learning AI. | | | | |
| VP01 | Students and colleagues frequently commend me for my ability to effectively teach complex Information and Communication Technology (ICT) subjects. | | | | |
| VP02 | During my studies at the University/National College of Education, my peers and instructors recognized me as having a strong grasp of Information and Communication Technology (ICT) and/or Artificial Intelligence (AI). | | | | |
| SC01 | I possess the necessary intellectual stamina and dedication to continuously learn and update my knowledge of Artificial Intelligence (AI), allowing me to effectively teach this subject. | | | | |
| SC02 | Even with the extra preparation involved, teaching a demanding subject like Artificial Intelligence (AI) doesn't necessarily stress or discourage me | | | | |
| IM01 | I aspire to become a highly regarded educator, recognized for my expertise in Artificial Intelligence (AI) and its potential impact on the future. | | | | |
| IM02 | Witnessing my students achieve success in the Information Technology (IT) sector, both locally and internationally, is a significant source of satisfaction for me as a teacher. | | | | |
| EX01 | Limited resources like computers, internet access, and labs at my school don't significantly hinder my ability to effectively teach Artificial Intelligence. | | | | |
| EX02 | I am confident that the school administration would be supportive in addressing any challenges that may arise while teaching Artificial Intelligence (AI). | | | | |
| EX03 | I believe my students' existing knowledge, computational skills, critical thinking abilities, and everyday experiences provide a strong foundation for understanding Artificial Intelligence (AI). | | | | |
| EX04 | I am confident the government will offer us comprehensive and adequate training opportunities for teaching Artificial Intelligence (AI). | | | | |

To gain a deeper understanding, the individual responses for the mastery experience indicator variables were further analysed. Figure 4.2 highlights nine questions in red where over 50% of respondents chose answers (a) or (b). Referring to Table 3.2, this suggests that a majority of ICT teachers: 1) Haven't formally studied artificial intelligence (AI) at the tertiary level, 2) Cannot identify common AI applications and their underlying technologies, 3) Have little to no practical experience with AI, 4) Don't engage in continuous learning about AI, 5) Lack interaction with peers or students to develop AI skills, 6) Don't possess the self-motivation or confidence to acquire new AI knowledge, share it, or guide students in AI innovation, and 7) Have no social capital to stay informed about industry AI requirements. However, four other questions show that over 50% of respondents chose answers (c) or (d). This indicates that a majority of respondents: 1) Believe they can overcome the challenge of teaching AI similarly to past experiences with emerging technologies, 2) Actively seek new tools and teaching methods for effective AI instruction, 3) Project confidence as ICT teachers in the classroom, and 4) Are proficient in preparing students for exams. Overall, these results suggest limited mastery experience with AI among respondents. They believe they can adapt existing teaching methods for other ICT topics to AI, which might be insufficient. Teaching a complex subject like AI requires a deeper understanding and a different approach compared to traditional exam preparation methods to equip young learners with the necessary AI skills.

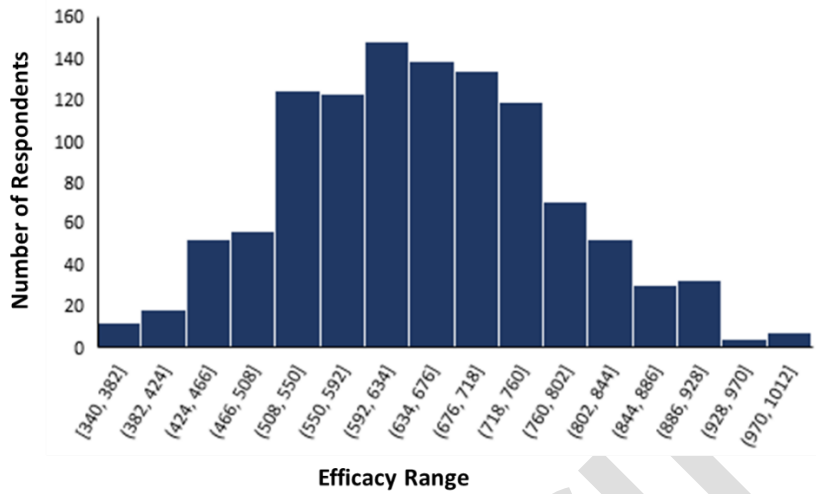

**Figure 4.1:** Distribution of individuals' overall self-efficacy scores

|     | MS01 | MS02 | MS03 | MS04 | MS05 | MS06 | MS07 | MS08 | MS09 | MS10 | MS11 | MS12 | MS13 |
|-----|------|------|------|------|------|------|------|------|------|------|------|------|------|
| (a) | 43%  | 12%  | 44%  | 7%   | 26%  | 30%  | 4%   | 2%   | 1%   | 26%  | 21%  | 70%  | 74%  |
| (b) | 31%  | 49%  | 41%  | 27%  | 50%  | 49%  | 31%  | 40%  | 9%   | 29%  | 45%  | 20%  | 20%  |
| (c) | 21%  | 30%  | 12%  | 39%  | 17%  | 16%  | 37%  | 24%  | 45%  | 23%  | 22%  | 7%   | 4%   |
| (d) | 5%   | 10%  | 3%   | 26%  | 8%   | 5%   | 28%  | 35%  | 45%  | 22%  | 12%  | 3%   | 2%   |

**Figure 4.2:** Response values of the indicator variables of Mastery Experience

The responses for other self-efficacy indicators, including Vicarious Experiences, Verbal Persuasion, Emotional and Physiological Factors, Imaginary Experiences, and External Factors, were also analysed (Figure 4.3). Notably, over half of the respondents chose answers (a) or (b) for Vicarious Experiences and Verbal Persuasion. This suggests:

- A potential lack of motivation within peer groups - They may not be witnessing high-quality teaching from colleagues.
- A serious lack of mentor-mentee relationships - Over 80% reported lacking a proper mentor to guide their teaching development.
- Little to no experience of receiving appreciation for successful teaching.

However, the responses for Emotional and Physiological Factors, Imaginary Experiences, and External Factors differ significantly from Vicarious Experiences and Verbal Persuasion. Over half of the respondents chose answers (c) or (d) for these indicators. This suggests that the Emotional and Physiological Factors, and Imaginary Experiences might be influenced by the high cultural value placed on the teaching profession in Sri Lanka. Respondents might be responding based on their perception of the ideal teacher role rather than their actual experiences. The extremely low percentage of respondents choosing answer (a) for these questions strengthens this argument.

|   | VE01 | VE02 | VE03 | VP01 | VP02 | EP01 | EP02 | IE01 | IE02 | EX01 | EX02 | EX03 | EX04 |
|---|------|------|------|------|------|------|------|------|------|------|------|------|------|
| (a) | 18% | 11% | 44% | 16% | 46% | 3% | 2% | 3% | 1% | 15% | 8% | 10% | 4% |
| (b) | 44% | 47% | 41% | 41% | 33% | 20% | 17% | 17% | 7% | 33% | 34% | 43% | 16% |
| (c) | 25% | 29% | 10% | 33% | 15% | 32% | 45% | 29% | 18% | 28% | 32% | 32% | 23% |
| (d) | 14% | 14% | 5% | 10% | 6% | 45% | 36% | 51% | 74% | 24% | 26% | 15% | 57% |

**Figure 4.3:** Response values of the indicator variables of Vicarious Experiences, Verbal Persuasion, Emotional and Physiological Factors, Imaginary Experiences, and External Factors

## 4.2 Analysis of the Conceptual Model

This section analyses the conceptual model of self-efficacy presented in Section 2.2 using Partial Least Squares Structural Equation Modeling (PLS-SEM) [62]. The goal is to gain a deeper understanding of the factors influencing teacher self-efficacy in teaching AI. The inner model includes six latent independent variables (exogenous constructs): Mastery Experiences, Vicarious Experiences, Verbal Persuasion, Emotional and Physiological Factors, Imaginary Experiences, and External Factors. These constructs are believed to influence a single latent dependent variable (endogenous construct): Self-Efficacy. The outer model indicators were primarily developed from the questions in the efficacy scale explained in Section 2.2. In simpler terms, each question on the scale corresponds to a single indicator variable for a specific latent variable in the inner model. A critical question was whether these indicator variables were formative or reflective. The key distinction lies in interchangeability. Formative indicators contribute to building the latent variable and should not be interchangeable [63]. Conversely, reflective indicators represent different aspects of the latent variable and can potentially be substituted. Based on the nature of the indicators and their interchangeability, all latent constructs in the model were determined to be reflective. The PLS-SEM model was implemented using SmartPLS 4.0 software. Figure 4.4 depicts the final model developed in SmartPLS.

The analysis resulted in removing some indicator variables from the "Mastery Experience" construct. Decisions about removing indicator variables were based on two criteria: the average variance extracted (AVE) value and the outer loading values of all indicators for that construct [64]. Following this approach, indicator variables MS01, MS08, MS09, MS10, MS12, and MS13 were excluded from the "Mastery Experience" construct. While a minimum outer loading of 0.7 is generally recommended for retaining indicators in reflective models [64], some exceptions were made for "Mastery Experience" and "External Factors." In these cases, certain indicators with outer loadings below 0.7 were retained based on their impact on the overall AVE of the construct, as suggested in [64]. Furthermore, the p-values associated with the relationships between each indicator variable and its corresponding latent construct all indicate strong positive relationships. The following subsections will look deeper into evaluating the measurement and structural models.

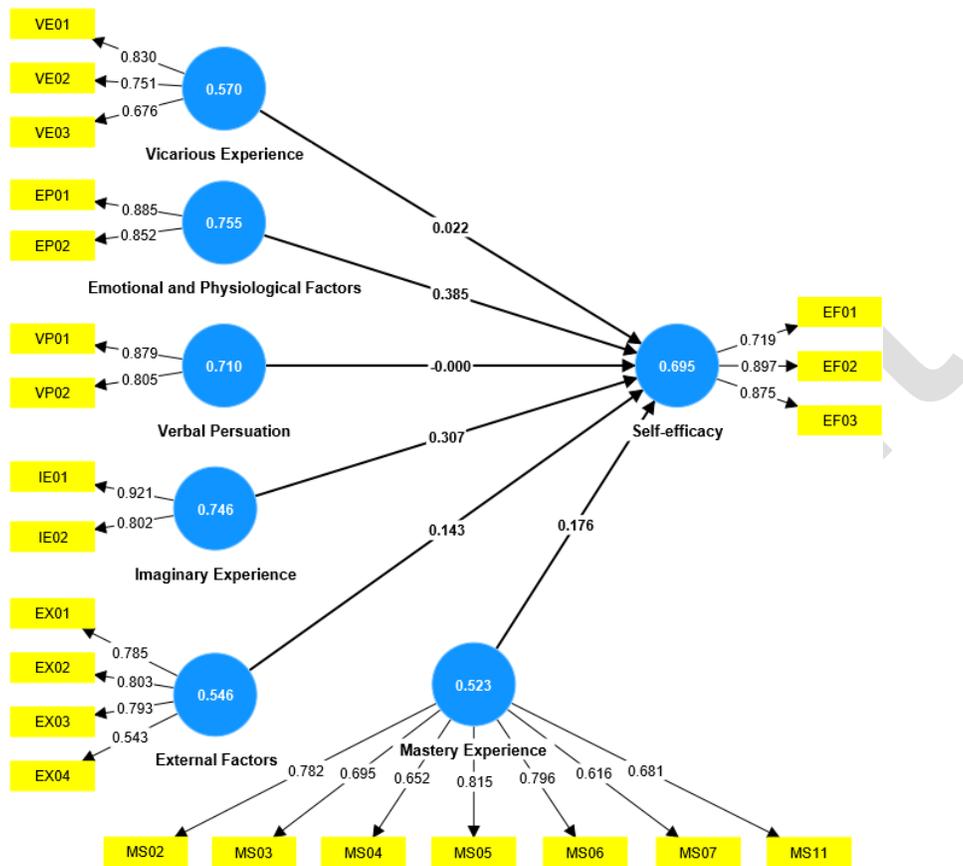

**Figure 4.4:** The PLS-SEM Model

## 4.3 Evaluation of the Measurement Model

The measurement model was evaluated using three key criteria: internal consistency reliability, convergent validity, and discriminant validity [64], [65]. Internal consistency reliability, assessed by Cronbach's Alpha and Composite Reliability (rho_c) [64], measures how well the indicators within a latent variable capture the same underlying concept [66]. As shown in Table 4.1, all latent variables have rho_c values exceeding 0.7, indicating good internal consistency. Minor discrepancies were observed for three latent variables with Cronbach's Alpha values slightly below 0.7. However, considering their high rho_c values, these deviations were deemed acceptable [67]. Convergent validity, evaluated using Average Variance Extracted (AVE) [64], examines whether theoretically related constructs are indeed related in the model [68]. AVE represents the average amount of variance in the indicators reflecting the latent variable. According to [64], an AVE value above 0.5 indicates adequate convergent validity. Table 4.1 demonstrates that all reflective latent variables have AVE values exceeding 0.5, suggesting good convergent validity for the measurement model. In conclusion, the evaluation using internal consistency reliability and convergent validity demonstrates that the reflective measurement model possesses strong measurement properties.

**Table 4.1: Evaluation of internal consistency reliability**

|  | Cronbach's alpha | Composite reliability (rho_c) | Average variance extracted (AVE) |
|---|---|---|---|
| **Emotional and Physiological Factors** | 0.676 | 0.860 | 0.755 |
| **External Factors** | 0.714 | 0.825 | 0.546 |
| **Imaginary Experience** | 0.672 | 0.854 | 0.746 |
| **Mastery Experience** | 0.845 | 0.884 | 0.523 |
| **Self-efficacy** | 0.778 | 0.872 | 0.695 |
| **Verbal Persuasion** | 0.595 | 0.830 | 0.710 |
| **Vicarious Experience** | 0.622 | 0.798 | 0.570 |

The final quality criterion assessed for the measurement model was discriminant validity, which examines whether unrelated constructs are indeed distinct in the model [69]. Discriminant validity was evaluated using two approaches: the Fornell-Larcker criterion and the Heterotrait-monotrait (HTMT) ratio. The Fornell-Larcker criterion, implemented in SmartPLS software, showed that all reflective latent variables have higher discriminant validity, as presented in Table 4.2. This result was further supported by observations from the cross-loadings matrix. However, the HTMT criterion identified one instance where the HTMT value exceeded 0.9, suggesting a potential lack of discriminant validity for that specific pair of constructs. Despite this, considering the overall positive results from the Fornell-Larcker test and the cross-loadings, it was concluded that the model possesses sufficient discriminant validity to proceed with further analysis [64].

**Table 4.2: Results of the Farnell-Larcker test for discriminant validity**

|  | Emotional and Physiological Factors | External Factors | Imaginary Experience | Mastery Experience | Self-efficacy | Verbal Persuasion | Vicarious Experience |
|---|---|---|---|---|---|---|---|
| **Emotional and Physiological Factors** | 0.869 |  |  |  |  |  |  |
| **External Factors** | 0.523 | 0.739 |  |  |  |  |  |
| **Imaginary Experience** | 0.572 | 0.493 | 0.864 |  |  |  |  |
| **Mastery Experience** | 0.501 | 0.434 | 0.397 | 0.723 |  |  |  |
| **Self-efficacy** | 0.733 | 0.581 | 0.676 | 0.566 | 0.834 |  |  |
| **Verbal Persuasion** | 0.358 | 0.322 | 0.288 | 0.559 | 0.382 | 0.842 |  |
| **Vicarious Experience** | 0.442 | 0.395 | 0.375 | 0.597 | 0.47 | 0.483 | 0.755 |

## 4.4 Evaluation of the Structural Model

The structural model's effectiveness was assessed using four key criteria: coefficient of determination ($R^2$), predictive relevance ($Q^2$), effect sizes ($f^2$), and the significance and strength of path coefficients [67]. $R^2$, as shown in Table 4.3, indicates the proportion of variance in the dependent variable ("Self-Efficacy") explained by the independent variables in the model. While minimum acceptable $R^2$ values vary by research field, social science studies generally consider values above 0.5 to be acceptable [70]. In this case, the $R^2$ value of 0.683 suggests a strong correlation between the independent variables and self-efficacy. Predictive relevance, measured by $Q^2$, goes beyond simply explaining variance. It assesses how well the model predicts the dependent variable based on the independent variables. As seen in Table 4.3, the $Q^2$ value for self-efficacy is well above zero, indicating the model has very high predictive power. This evaluation using $R^2$ and $Q^2$ suggests that the structural model effectively captures the relationships between the independent variables and self-efficacy in teaching AI.

**Table 4.3:** The $R^2$ and $Q^2$ values of the structural model

|  | R-square | R-square adjusted | $Q^2$predict | RMSE | MAE |
| --- | --- | --- | --- | --- | --- |
| Self-efficacy | 0.683 | 0.682 | 0.679 | 0.568 | 0.436 |

The strength and significance of the relationships between self-efficacy and the exogenous variables were evaluated using path coefficients and t-values [67]. Hair et al. [71] suggest that path coefficient values close to 0.5 or greater indicate large effects, values around 0.3 indicate medium effects and values below 0.1 indicate small effects. Additionally, for a relationship to be considered statistically significant at the 95% confidence level, a t-value above 1.96 and a p-value less than 0.05 are typically expected. Table 4.4 summarises these statistics for the relationships between the six exogenous variables and self-efficacy. As the table shows, emotional and physiological factors and imaginary experiences have path coefficients exceeding 0.3 with self-efficacy. These relationships are also statistically significant, with t-values well above 1.96 and p-values well below 0.05. This suggests that these factors have a strong and significant positive influence on self-efficacy. Mastery experience and external factors also have significant relationships with self-efficacy, although the path coefficients are weaker (below 0.3). However, verbal persuasion and vicarious experiences appear to have no statistically significant influence on self-efficacy.

**Table 4.4:** Size and significance of path coefficients

|  | Path coefficient | Sample mean (M) | Standard deviation | T statistics | P values |
| --- | --- | --- | --- | --- | --- |
| Emotional and Physiological Factors -> Self-efficacy | 0.385 | 0.384 | 0.027 | 14.51 | 0.0000 |
| External Factors -> Self-efficacy | 0.143 | 0.143 | 0.022 | 6.50 | 0.0000 |
| Imaginary Experience -> Self-efficacy | 0.307 | 0.307 | 0.024 | 12.76 | 0.0000 |
| Mastery Experience -> Self-efficacy | 0.176 | 0.176 | 0.024 | 7.48 | 0.0000 |
| Verbal Persuasion -> Self-efficacy | 0.000 | 0.001 | 0.019 | 0.00 | 0.9990 |
| Vicarious Experience -> Self-efficacy | 0.022 | 0.022 | 0.022 | 1.02 | 0.3060 |

To further analyse the influence of each exogenous variable on self-efficacy, the effect size measure, $f^2$, was employed [67]. This value essentially assesses the impact of removing a particular exogenous variable from the model on the endogenous variable (Self-efficacy). Hair et al. [67] suggest an $f^2$ value greater than or equal to 0.02, 0.15, and 0.35 indicates low, medium, and high effect sizes, respectively. As shown in Table 4.5, "Emotional and Physiological Factors" and "Imaginary Experience" have a moderate impact on self-efficacy, with emotional factors having a slightly stronger influence. "Mastery Experience" and "External Factors" exhibit a weak impact, while "Verbal Persuasion" and "Vicarious Experience" appear to have a negligible effect on self-efficacy. These findings suggest that emotional well-being and imaginary experiences play a more significant role in shaping self-efficacy for teaching AI than the other factors examined.

Table 4.5: The F² values of the structural model

|  | f-square |
|---|---|
| Emotional and Physiological Factors -> Self-efficacy | 0.252 |
| External Factors -> Self-efficacy | 0.041 |
| Imaginary Experience -> Self-efficacy | 0.182 |
| Mastery Experience -> Self-efficacy | 0.049 |
| Verbal Persuasion -> Self-efficacy | 0.000 |
| Vicarious Experience -> Self-efficacy | 0.001 |

## 4.5 Examining the Teachers' Level of Expertise in AI

Given the weak but significant relationship between mastery experience and self-efficacy identified earlier, a further analysis was conducted to explore teachers' AI expertise in more detail. The questionnaire included a section where participants ranked their knowledge of popular AI topics using a five-point scale: 0 (no knowledge), 1 (have heard of), 2 (average), 4 (good), and 5 (very good). Figure 4.5 summarises the findings. As the figure shows, the teachers' overall AI expertise appears to be quite low. Most respondents indicated either "no knowledge" or "have heard of" for all the listed AI topics. However, a few participants demonstrated somewhat satisfactory mastery in areas like Intelligent Agents, Machine Learning, Multi-Agent Systems, Genetic Algorithms, Artificial Neural Networks, and Ethics and Social Impact of AI. Unfortunately, this group represents a minority compared to the majority of teachers who have little to no experience with AI.

## 5 Discussion

The findings in Section 4.5 directly address the key research question of this paper: how prepared is the existing K-12 teacher workforce to deliver an AI curriculum? It is clearly evident from the results that the current level of AI expertise among teachers is largely insufficient. The AI4K12 policy framework outlines five core concepts for an AI curriculum: Perception, Representation & Reasoning, Learning, Natural Interaction, and Societal Impact. Based on this study's findings, it is evident that only a small percentage of teachers possess the competency to adequately cover content aligned with these AI4K12 guidelines. This conclusion is further reinforced by the average overall self-efficacy score presented in Section 4.1. These results reveal a significant lack of readiness among current teachers to effectively teach artificial intelligence. The next crucial question, then, is how to prepare teachers for the proposed curriculum if they are not currently ready. The PLS model presented in this paper offers valuable insights to address this challenge.

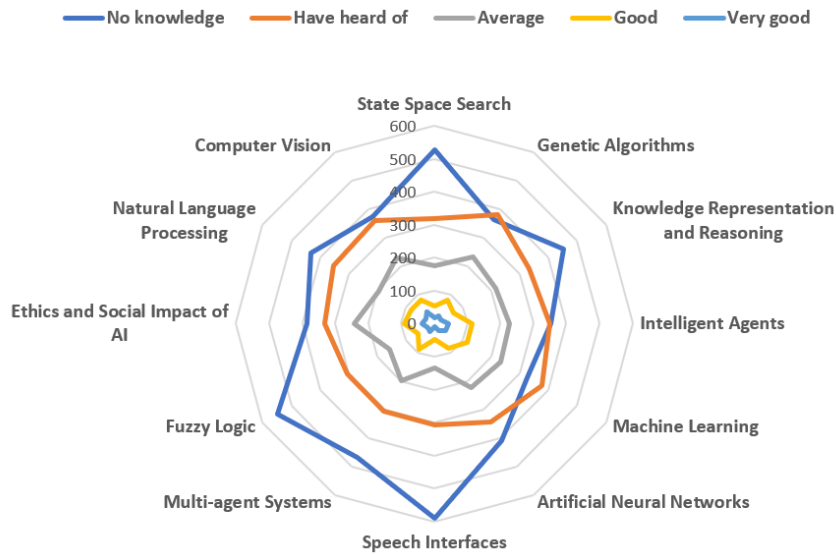

**Figure 4.5:** The mastery level of teachers on different AI techniques

## 5.1 Insights from the Structural Model

The study's findings offer valuable insights for designing professional development programs to enhance teachers' AI teaching competency. As revealed by the structural model evaluation in Section 4.4, vicarious experiences and verbal persuasion have no significant influence on teachers' self-efficacy. This lack of contribution suggests two key areas for improvement within teacher development programs. Firstly, the insignificant effect of vicarious experiences highlights a potential lack of role model teachers and mentors in the current system. Exposing teachers to successful AI educators through workshops, online resources, or peer observation programs could be beneficial. Secondly, the minimal impact of verbal persuasion suggests that teachers may not receive sufficient recognition or encouragement for developing AI teaching skills. Teacher professional development programmes could emphasise positive reinforcement through various strategies such as praising the efforts of teacher learners, constructive feedback mechanisms during learning, encouraging them to take risks and innovate when learning to teach, and rewarding them for teaching excellence.

The structural model revealed another interesting finding: emotional and physiological factors, along with imaginary experiences, had a greater contribution to teachers' self-efficacy for teaching AI than mastery experiences. Given the limited integration of AI into the current ICT curriculum, this could suggest that teachers are drawing on their past experience teaching other ICT topics, which were also challenging. The higher contribution of imaginary experiences might be culturally specific to Sri Lanka, where teaching is a respected profession. This could explain why teachers are optimistic about their success in teaching AI despite the apparent lack of mastery. Interestingly, while mastery experience has a weak but significant positive impact on self-efficacy, a closer look reveals this likely stems from specific survey items (MS04 and MS07). These items appear to tap into the respondents' general belief that they can handle teaching AI similarly to their past experiences with emerging technologies. Overall, these findings suggest a complex interplay of factors influencing teacher self-efficacy for AI education. While confidence in existing teaching skills is positive, it highlights the potential need for targeted professional development programs that address the specific challenges and opportunities of integrating AI into the curriculum.

The study's findings can be further explored through the lens of employability capital, a widely used concept in job market research [72]. This model identifies five key facets that contribute to a graduate's employability: human capital (skills and knowledge), social capital (networks and relationships), cultural capital (cultural awareness and competence), identity capital (self-perception and career aspirations), and psychological capital (resilience and adaptability). This framework has been extended to include agentic capital, which emphasises an individual's initiative and proactiveness [73]. These various forms of capital collectively influence a person's chances of obtaining and retaining employment. In the context of this research, adapting to the requirement of teaching artificial intelligence is analogous to an employability challenge for ICT teachers. Therefore, their self-efficacy for teaching AI (or any subject) can be seen as a reflection of their overall employability capital. The low self-efficacy levels observed in the study suggest a potential lack of employability capital among the teachers. Psychological capital, defined as the ability to overcome barriers, adapt to new situations, and respond proactively to career challenges, is a key driver of resilience and self-efficacy [72]. Interestingly, the analyses in Figures 4.2 and 4.3 reveal that the respondents tended to agree with survey questions that indicate higher levels of psychological capital. For example, over 60% of respondents indicated successful past experiences in overcoming challenges related to learning and teaching new technologies. This positive outlook aligns with the conceptual model's prediction that psychological capital contributes to self-efficacy. Moreover, it explains why the emotional and physiological factors have a higher contribution to overall self-efficacy: a majority believe that they can adapt to teaching AI just as they have adapted to teaching other emerging technologies in the past. Conversely, the analysis in Figures 4.2 and 4.3 suggests lower levels of human, social, and cultural capital among the teachers. These deficiencies likely affect their mastery experiences, vicarious experiences, and even susceptibility to verbal persuasion. These critical findings highlight the need for targeted interventions to address these specific aspects of employability capital and ultimately enhance teachers' self-efficacy for AI education. The critical findings of the research could be summarised as follows.

- The readiness of the current teacher workforce to teach artificial intelligence seems low
- The teachers' knowledge and skills related to AI is alarmingly insufficient
- The respondents' self-efficacy in teaching AI mainly stems from the belief that they can handle the challenge of teaching AI similarly to their past experiences with emerging technologies.
- Important sources of self-efficacy, such as vicarious experience and verbal persuasion, are currently insignificant in the context of Sri Lankan ICT teachers
- The respondents exhibit minimal social and cultural capital to succeed in the given endeavour

## 5.2 Implications of the findings

The second research question focused on strategies to prepare teachers for the proposed AI curriculum. The study's findings offer valuable insights for policymakers addressing this challenge. Currently, the Ministry of Education and the National Institute of Education primarily rely on a direct, face-to-face teacher training approach. However, resource limitations hinder island-wide coverage, prompting a recent shift towards the "train-the-trainer" model. Additionally, efforts have been made to incorporate learner-centered approaches. However, the study raises questions about whether these existing methods can effectively enhance teachers' self-efficacy for teaching advanced subjects like AI, particularly in engaging young learners. The low contribution of vicarious experiences and verbal persuasion to self-efficacy suggests a need for a more systemic approach. This research recommends a connected teacher professional development system, fostering stronger and more frequent connections among teachers and stakeholders with valuable resources. As connectivism learning theory suggests [74], continuous and evolving networks on technology-assisted platforms enable the exchange of knowledge, ideas, and resources. Building on this concept, the research proposes a socio-technical system for teacher development, supported by external stakeholders, as an alternative to the existing approach. This approach aligns with the ecological perspective of learning [75] and could potentially alleviate resource constraints. However, further research is needed to determine the likely success and sustainability of such a complex learning ecosystem for teachers.

## 5.3 Limitations and Future Work

This study has some limitations that offer opportunities for future research. While the self-efficacy scale primarily drew upon existing literature, some items lacked formal reliability and validity testing. Future research can address this by developing a rigorously validated and reliable scale specifically designed to assess AI teachers' self-efficacy. The analysis did not consider factors such as age, gender, geographical location, or educational background. For example, younger teachers might exhibit higher self-efficacy and be more connected with others in the educational system than older teachers. Additionally, teachers with a background in mathematics (from GCE Advanced Level exams and bachelor's degrees) might be better equipped to grasp AI concepts than those with social science or commerce backgrounds. These factors could potentially act as mediating variables in the proposed conceptual model. One future direction is to extend this research by analysing the impact of these mediating factors on teachers' self-efficacy for teaching AI. Another area for future exploration involves the capital-based view of teacher competence for AI education presented in the discussion. A deeper analysis of the collected data from a capital perspective (human, social, cultural, psychological, and agentic) could provide valuable insights for designing a socio-technical teacher training system. In essence, understanding the teachers' overall capital would further inform the development of effective interventions.

## 6 conclusion

This paper investigated Sri Lankan secondary and upper-secondary ICT teachers' self-efficacy for teaching artificial intelligence (AI). The goal was to assess their preparedness for the government's planned integration of AI into the 2024 academic year curriculum. The study provides compelling evidence that teachers' current self-efficacy and, consequently, their readiness to teach AI is very low. The data analysis revealed a critical lack of vicarious experiences (learning from others), verbal persuasion (encouragement), and mastery experiences (experiences of successful encounters of mastery). Drawing on a capital-based view of employability, the study concludes that teachers are not currently prepared to teach AI. This research is one of the first (if not the very first) in Sri Lanka to assess ICT teachers' readiness for an advanced technology like AI. The findings hold significant value for policymakers who are developing professional development programs to equip teachers for success in AI4K12 initiatives. Moreover, these results could inform AI4K12 policymakers and researchers worldwide, serving as a benchmark for strategising teacher development programs in their respective countries. The study recommends a systems thinking approach to teacher training to address self-efficacy issues. This approach opens doors for future research based on complex socio-technical systems.